\documentclass[conference]{IEEEtran}
\IEEEoverridecommandlockouts
\usepackage{cite}
\usepackage{amsmath,amssymb,amsfonts}
\usepackage{algorithmic}
\usepackage{graphicx}
\usepackage{textcomp}
\usepackage{xcolor}
\usepackage{graphicx}
\usepackage{subcaption}
\usepackage[normalem]{ulem}

\renewcommand{\figurename}{Figure}
\def\BibTeX{{\rm B\kern-.05em{\sc i\kern-.025em b}\kern-.08em
    T\kern-.1667em\lower.7ex\hbox{E}\kern-.125emX}}

\begin{document}

\title{E-TSL: A Continuous Educational Turkish Sign Language Dataset with Baseline Methods }

\author{\IEEEauthorblockN{Şükrü Öztürk}
\IEEEauthorblockA{\textit{Computer Engineering Department} \\
\textit{Hacettepe University}\\
Ankara, TÜRKİYE \\
sukruozturk@hacettepe.edu.tr}
\and
\IEEEauthorblockN{Hacer Yalim Keles}
\IEEEauthorblockA{\textit{Computer Engineering Department} \\
\textit{Hacettepe University}\\
Ankara, TÜRKİYE \\
hacerkeles@cs.hacettepe.edu.tr}
}

\maketitle

\begin{abstract}

This study introduces the continuous Educational Turkish Sign Language (E-TSL) dataset, collected from online Turkish language lessons for 5th, 6th, and 8th grades. The dataset comprises 1,410 videos totaling nearly 24 hours and includes performances from 11 signers. Turkish, an agglutinative language, poses unique challenges for sign language translation, particularly with a vocabulary where 64\% are singleton words and 85\% are rare words, appearing less than five times. We developed two baseline models to address these challenges: the Pose to Text Transformer (P2T-T) and the Graph Neural Network based Transformer (GNN-T) models. The GNN-T model achieved 22.93 ROUGE-L score, 21.01\% BLEU-1 score and 3.49\% BLEU-4 score, presenting a significant challenge compared to existing benchmarks. Additionally, we benchmarked our model using the well-known PHOENIX-Weather 2014T dataset to validate our approach.

\end{abstract}



\section{\textbf{Introduction}}

According to World Health Organization data\cite{who}, 432 million adults worldwide are deaf. Most people with congenital hearing impairments, in particular, try to meet their communication needs with sign language. Sign language is a communication tool that involves creating various and meaningful visual elements using various parts of the body, particularly the hands and face. In order to communicate with sign language, it is necessary to have mutual knowledge of the language. Sign languages of hearing-impaired individuals, like spoken language, differ from country to country. This makes it difficult for two hearing-impaired individuals, who are citizens of different countries, to communicate using sign language. They also have difficulty communicating with other people who do not know sign language.

Today, manufacturers produce a variety of devices to overcome hearing impairments. These devices allow you to hear the sounds and learn the spoken language, but their cost is very high. According to the World Health Organization, nearly 80\% of hearing-impaired individuals live in low and middle-income countries. For this reason, most hearing-impaired people cannot access these devices. Fortunately, people who do not have access to these devices are not without alternatives, thanks to the incredible developments in computer vision, sign language recognition, and translation in recent years.

The first attempts to recognize sign language date back to the 1970s, when researchers used robotic hands or gloves to capture finger-spelling or isolated signs \cite{parton}. Later, more advanced techniques are developed to recognize isolated signs \cite{Tur19}, \cite{Tur21}, and continuous signs or sentences using computer vision methods, such as Hidden Markov Models or neural networks \cite{bragg}. However, sign language recognition is still a challenging problem due to the complexity and variability of sign languages, which involve not only hand shapes and movements but also facial expressions, body posture, and context.

Sign language generation is the process of automatically producing signs based on a given input, such as text or speech. The first attempts at generating sign language used avatars or anime characters to display finger-spelling or isolated signs \cite{huenerfauth}. Later, more sophisticated systems were developed to generate continuous signs or sentences using natural language processing methods, such as syntactic analysis or semantic representation \cite{kipp}. However, sign language generation is also a challenging problem due to the need to preserve the grammatical and pragmatic features of sign languages, as well as the naturalness and expressiveness of the signs.

Sign language translation is the task of automatically converting between sign language and spoken or written language. The first attempts to translate sign language used rule-based methods or statistical models to map between signs and words \cite{morissey}. Later, more advanced systems were developed to translate between sign language and spoken or written language using neural machine translation methods, which can learn from large parallel corpora of sign language videos and text or speech transcripts \cite{camgoz2019}. With the article "Attention is all you need" \cite{vaswani} published in 2017, significant progress has been made in sign language translation as well as in natural language processing with transformers\cite{camgoz2020}.

Although large-scale Turkish Sign Language datasets and related research exist in isolated domains \cite{Sincan20}, \cite{Sincan22}, \cite{Bosphorussign22k}, there is currently no continuous TSL dataset in the literature. In this study, we introduce a new Educational Turkish Sign Language (E-TSL) dataset and propose two transformer based deep models that are trained using this dataset, serving as baseline models for this dataset for future research. The paper is organized as follows: Section II presents the related works; Section III presents material and methods; Section IV presents experimental results and discussion; while Section V concludes the study.

\section{\textbf{Related Works}}

Researchers have conducted several studies in the field of sign language recognition and translation. In this section, we present a comprehensive review of some important works in this domain.

Gokul et al.\cite{gokul} have published a 4.6K hour dataset of 10 different sign languages, which they call SignCorpus. They also published the graph-based Sign2Vec model they trained with this dataset. They created MultiSign-ISLR –a multilingual and label- aligned dataset of sequences of pose keypoints from 11 labelled datasets across 7 sign languages, and MultiSign-FS – a new finger-spelling training and test set across 7 languages. As a result, they worked on 7 different datasets and presented models that could be fine tuned and used in multilingual sign language translations.

Bull et al.\cite{bull} studied the alignment of asynchronous subtitles in sign language videos. They propose a Transformer architecture, which is trained on manually annotated alignments covering over 15K subtitles that span 17.7 hours of video. The two inputs are encoded using BERT subtitle embeddings and CNN (Convolutional Neural Network) video representations that have been trained to recognise signs. These inputs interact with each other through a series of attention layers.

Chen et al.\cite{chen} introduce a dual visual encoder containing two separate streams to model both the raw videos and the keypoint sequences generated by an off-the-shelf keypoint estimator. To make the two streams interact with each other, they explore a variety of techniques, including bidirectional lateral connection, sign pyramid network with auxiliary supervision, and frame-level self-distillation. TwoStream-SLR, the resulting model, demonstrates competence in sign language recognition (SLR). They used PHOENIX14\textbf{T} and CSL Daily datasets and achieved ROUGE: 51.59\%, BLEU-4: 26.71\% using Sign2Gloss2Text and ROUGE: 53.48\%, BLEU-4: 28.95\% using Sign2Text on PHOENIX14\textbf{T} and achieved ROUGE: 54.92\%, BLEU-4: 24.13\% using Sign2Gloss2Text and ROUGE: 55.72\%, BLEU-4: 25.79\% using Sign2Text on CSL Daily dataset.

Saunders et al.\cite{saunders} proposed Progressive Transformers, the first SLP (Sign Language Production) model to translate from discrete spoken language sentences to continuous 3D sign pose sequences from beginning to end. Transformer architecture they proposed can translate from discrete spoken language to continuous sign pose sequences.

Camgöz et al.\cite{camgoz2020} proposed a transformer-based architecture that jointly learns Continuous Sign Language Recognition and Translation and can be trained end-to-end. They used PHOENIX14\textbf{T} dataset to evaluate their architecture. They achieved 24.49\% WER and 21.80 BLEU-4 score for Sign2(Gloss+Text) on Sign Language Recognition and 26.16\% WER and 21.32 BLEU-4 score for Sign2(Gloss+Text) on Sign Language Translation.

Chen et al.\cite{chen2022} proposed that in order to overcome the bottleneck problem of data in the training of sign language translation models, they suggest pre-training the model in the general domain first and then pre-training it gradually with the within-domain dataset. In addition to pre-training, they increased the performance even more by applying the transfer learning method. They used PHOENIX14\textbf{T} and CSL Daily datasets and achieved ROUGE: 51.43\%, BLEU-4: 21.46\% using Sign2Gloss2Text and ROUGE: 53.25\%, BLEU-4: 23.92\% using Sign2Text on CSL Daily and achieved ROUGE: 52.64\%, BLEU-4: 28.39\% using Sign2Text on PHOENIX14\textbf{T} dataset.

Cheng et al.\cite{cheng} proposed a fully convolutional network (FCN) for online SLR, aiming to simultaneously acquire spatial and temporal features from weakly annotated video sequences, where only sentence-level annotations are provided. The network incorporates a gloss feature enhancement (GFE) module to enhance sequence alignment learning. Remarkably, the proposed network can be trained end-to-end without the need for any pre-training. With the FCN they suggested, they obtained a word error rate (WER) of 3\% on the CSL dataset, and 23.9\% WER on the RWTH dataset.

Hao et al.\cite{hao} proposed a Self-Mutual Knowledge Distillation (SMKD) method, which enforces the visual and contextual modules to focus on short-term and long-term information while simultaneously increasing the discriminative power of both modules. Specifically, the visual and contextual modules share the weights of their corresponding classifiers and train with CTC (Connectionist Temporal Classification) loss simultaneously. Moreover, the spike phenomenon is widely associated with CTC loss. With this proposed method, they obtained 21\% WER on the PHOENIX14 dataset and 22.4\% WER on the PHOENIX14\textbf{T} dataset.

Hosain et al.\cite{hosain} proposed a method for word-level sign recognition from American Sign Language (ASL) using video. Their method employs both motion and hand shape cues, while remaining robust to variations in execution. They exploit the knowledge of the body pose, estimated by an off-the-shelf pose estimator. They pool spatio-temporal feature maps from different layers of a 3D convolutional neural network, using the pose as a guide. They train separate classifiers using pose-guided pooled features from different resolutions and fuse their prediction scores during test time. The proposed approach achieves 10\%, 12\%, 9.5\% and 6.5\% performance gains on WLASL100, WLASL300, WLASL1000, WLASL2000 subsets, respectively.

\begin{table*}[htbp]
\centering
\caption{Detailed Statistics of Prominent Continuous Sign Language Datasets}
\begin{tabular}{ p{2cm}||p{1.5cm}||p{2.25cm}|p{2.8cm}|p{2cm}|p{2.75cm} }
 & E-TSL & PHOENIX14\textbf{T}\cite{camgoz2018} & SWISSTXT-NEWS\cite{camgoz2021} & VRT-NEWS\cite{camgoz2021} & Elementary23-SLT\cite{voskou} \\
 \hline
 Total Words & 169.356 & 99.081 & 72.892 & 79.833 & 83.327 \\
 Vocabulary Size & 6.980 & 2.287 & 10.561 & 6.875 & 8.202 \\
 Singletons & 4.466 (64\%) & 1.077 (47\%) & 5.969 (57\%) & 3.405 (50\%) & 3.327 (41\%) \\
 Rare Words \textless{} 5 & 5.936 (85\%) & 1.758 (77\%) & 8.779 (83\%) & 5.334 (78\%) & 6.155 (75\%) \\
\end{tabular}
\label{tab:dataset_stats}
\end{table*}

Kan et al.\cite{kan} proposed a hierarchical spatio-temporal graph neural network (HST-GNN) to process graph representations of sign language movements. The distinctive features of sign languages are represented as hierarchical spatio-temporal graph structures, encompassing both high-level and fine-level graphs. In these graphs, each vertex represents a specific body part, while the edges denote the interactions between these parts. The high-level graphs capture patterns within regions like the hands and face, while the fine-level graphs focus on the joints of hands and landmarks of facial regions. This hierarchical approach allows for a comprehensive representation of the unique characteristics found in sign languages. The language decoder receives the resulting graphs after applying the convolution and pooling operations, and processes them to convert them into text. After applying the convolution and pooling operations to the resulting graphs, these data are sent to the language decoder and converted to text after processing there. They achieved a BLEU-4 score of 22.3 on the PHOENIX14\textbf{T} dataset and 17.8 on the CSL dataset with this network.

These works demonstrate various approaches and methodologies employed in the field of sign language recognition and translation. They contribute to the development of robust models, datasets, and techniques that aim to reduce the communication gap between sign language users and non-users.

\section{\textbf{Material and Methods}}

\subsection{\textbf{Proposed E-TSL Dataset}}

\begin{figure}[htb]
    \includegraphics[width=.4\textwidth]{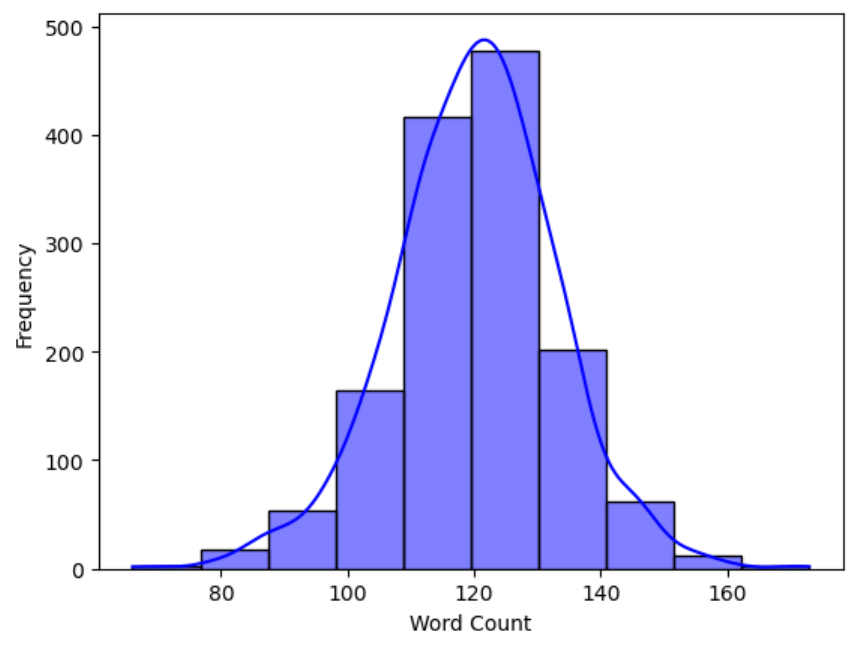}
    \caption{Word Distribution of E-TSL Dataset}
    \label{fig:word_distribution}
\end{figure}

The E-TSL dataset developed for this study comprises videos from 5th, 6th, and 8th grade Turkish language lessons gathered together from online lectures of TRT (Türkiye Radio Television Agency) EBA TV, along with their corresponding spoken language transcriptions. The dataset consists of 1,410 videos in total and is around 24 hours long. Videos are 190 x 230 pixels in size and contain 25 frames per second. We organized the videos in our dataset into small clips, each approximately 1 minute in length. Unlike existing datasets, which typically annotate videos at the sentence level, our clips provide longer continuous samples. Figure \ref{fig:word_distribution} shows the distribution of word counts per clip, depicted through a histogram with a normal distribution curve overlaid. This visualization demonstrates the typical clip length and the variability in text lengths within our dataset. Specifically, the word count per clip ranges from 80 to 160 words, with an average of approximately 120 words. There are 1,057 videos (75\%) for the train set, 141 videos (10\%) for the development set and 212 videos (15\%) for the test set.

In order to compare E-TSL with the 3 most used datasets as benchmarks in this domain, we present the data obtained on the basis of vocabulary-based measurement metrics in Table \ref{tab:dataset_stats}. To this comparison, we also include the recent Elementary23-SLT dataset, which has educational content similar to E-TSL. Considering that the PHOENIX14\textbf{T} dataset contains only weather news, its vocabulary size, singletons, and rare words are much smaller than the others. This ensures that PHOENIX14\textbf{T} gives better performance than others. The E-TSL dataset contains the largest number of words relative to its size compared to similar datasets. Although the vocabulary size of E-TSL is comparable to those of the SWISSTXT-NEWS, VRT-NEWS, and Elementary23-SLT datasets, it has a higher percentage of singletons and rare words (64\% and 85\%, respectively). This is primarily attributed to the agglutinative nature of the Turkish language, posing a challenge for models using the E-TSL dataset.

\subsection{\textbf{RWTH-PHOENIX-Weather 2014T (PHOENIX14\textbf{T}) Dataset}}

\begin{figure}[htb]
    \includegraphics[width=.08\textwidth]{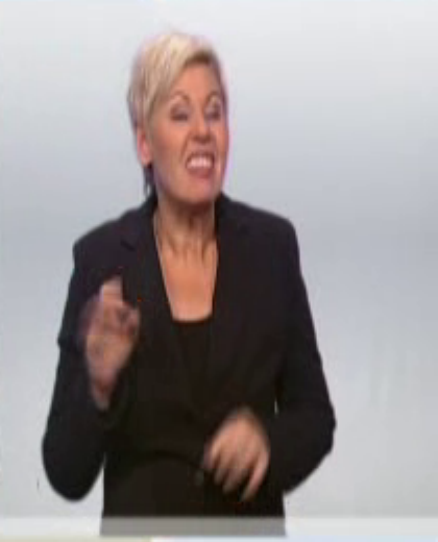}\hfill
    \includegraphics[width=.08\textwidth]{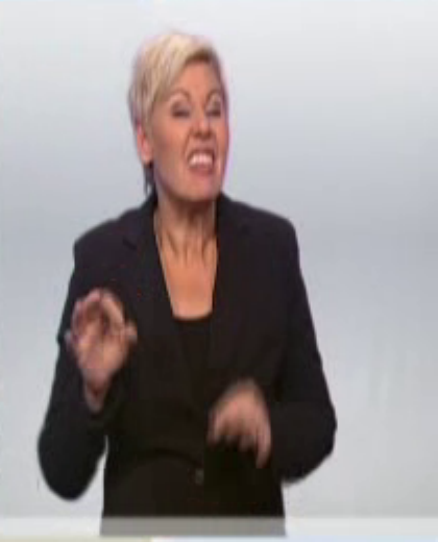}\hfill
    \includegraphics[width=.08\textwidth]{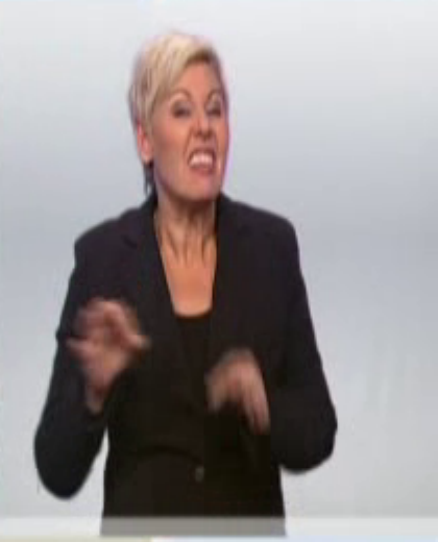}\hfill
    \includegraphics[width=.08\textwidth]{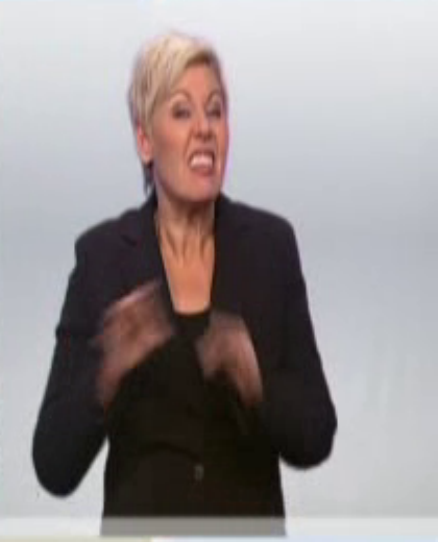}\hfill
    \includegraphics[width=.08\textwidth]{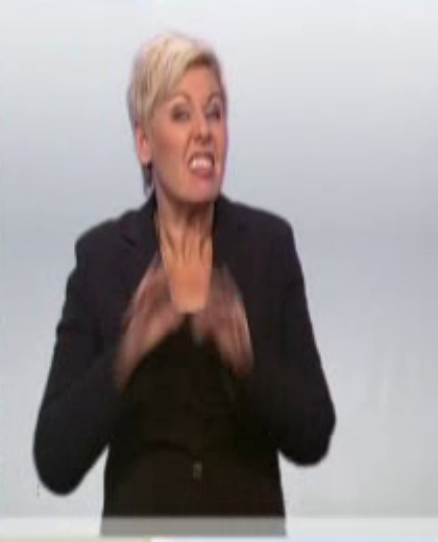}\hfill
    \includegraphics[width=.08\textwidth]{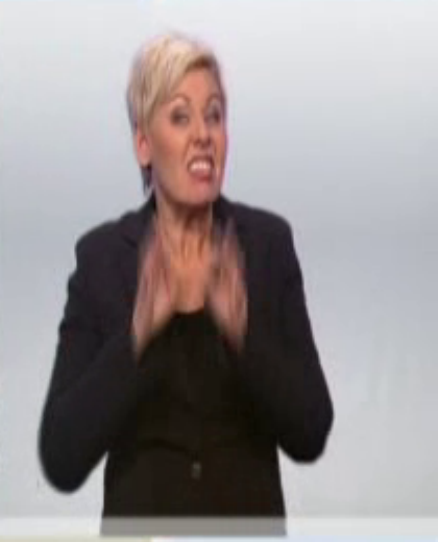}\hfill
    \\[\smallskipamount]
    \includegraphics[width=.08\textwidth]{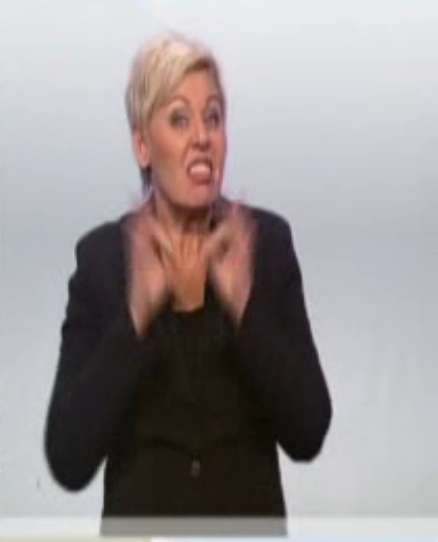}\hfill
    \includegraphics[width=.08\textwidth]{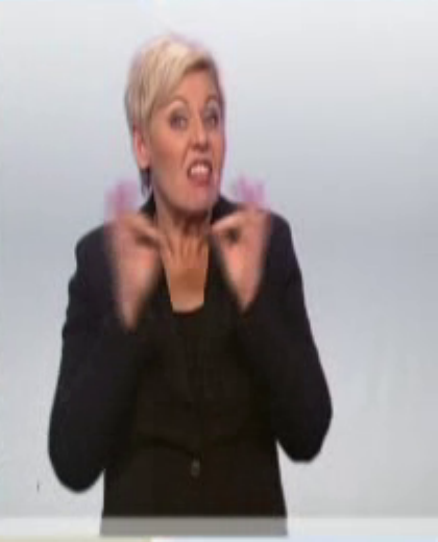}\hfill
    \includegraphics[width=.08\textwidth]{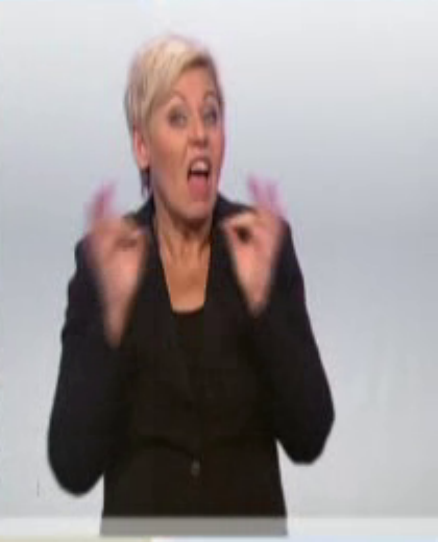}\hfill
    \includegraphics[width=.08\textwidth]{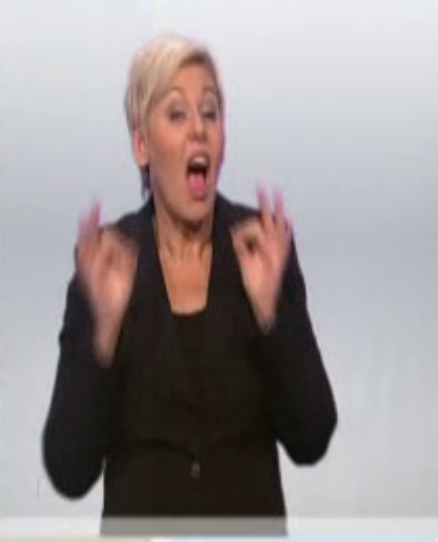}\hfill
    \includegraphics[width=.08\textwidth]{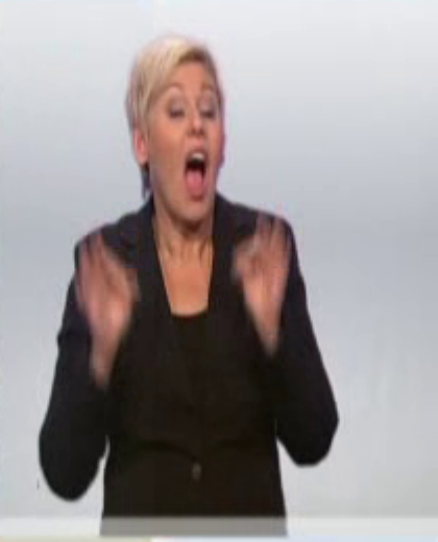}\hfill
    \includegraphics[width=.08\textwidth]{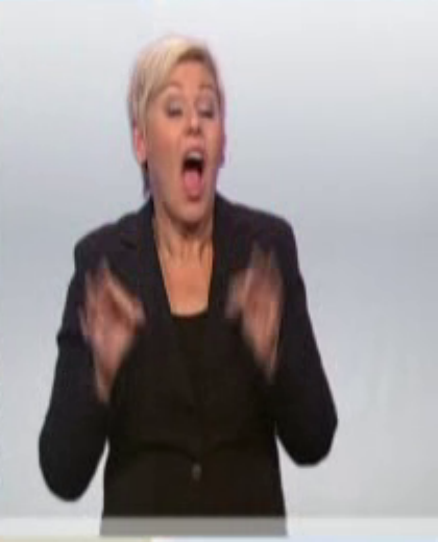}
    \caption{Sample Images from PHOENIX14\textbf{T} Dataset \cite{camgoz2018}}
    \label{fig:sampleImages}
\end{figure}

In this study, we also used the PHOENIX14\textbf{T} dataset\cite{camgoz2018}, which is widely used in the literature and has become a benchmark, in order to validate our models and make a performance comparison with E-TSL dataset. The PHOENIX14\textbf{T} dataset consists of sign language interpretation videos of weather forecasts on the German PHOENIX television channel. All videos are 25 frames per second, and the frame size is 210 x 260 pixels. Each frame shows the interpreter box only, as can be seen in Figure \ref{fig:sampleImages}. The videos in the dataset are divided into 8,257 parts, each corresponding to a sign language gloss and a reference sentence, and these parts are divided into frames. 7,096 of these parts are reserved for training, 519 for development, and 642 for testing. Each of these parts contains around 120 images on average.

\subsection{\textbf{Methodology}}

\begin{figure*}
\centerline{\includegraphics[width=\textwidth, keepaspectratio]{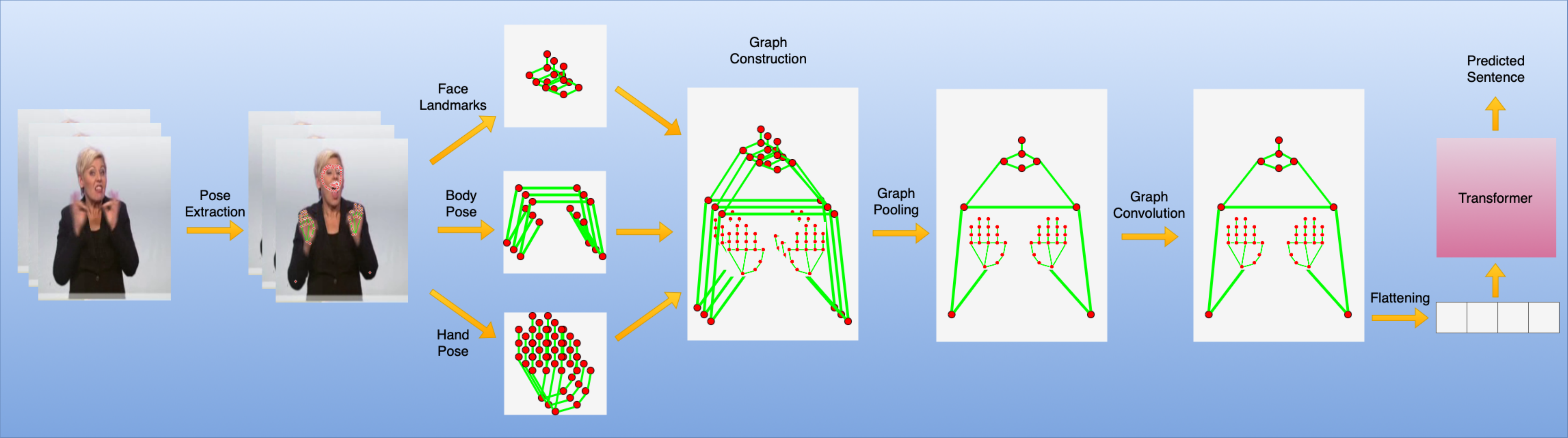}}
\caption{Architecture of the GNN-T Model}
\label{fig:architecture}
\end{figure*}

In this study, we developed two baseline models. The first one is a Graph Neural Network based Transformer (GNN-T) model, which extends the work of Kan et al.\cite{kan} by incorporating body pose data, recognizing the relevance of arm, hand, and facial movements in sign language. Our GNN-T model uses techniques like graph pooling and graph convolution, as shown in Figure \ref{fig:architecture}.

The second model is a Pose to Text Transformer (P2T-T), developed drawing inspiration from the work of Camgöz et al.\cite{camgoz2020}. We adapted this model to include pose modality and applied landmark normalization to enhance its performance. There is no gloss annotation in our model, hence it's trained end-to-end for translation; we modified the model to support Sign2Text by directly connecting the transformer encoder's output to the transformer decoder.

Below, we detail the key components of our models, outlining the technical aspects and functionalities that contribute to their performance.

\subsubsection{\textbf{Pose Extraction}}

In this study, we used the MediaPipe library\cite{lugaresi} to capture sign language movements and to represent the instantaneous positions of the hands, face, and certain parts of the body as graphs. Google researchers developed the MediaPipe framework, which detects various body parts from videos and images and presents them to the user as graphs. It offers a graph structure consisting of 468 points for the face, 33 points for the body, and 21 points for each hand. Head and lip movements are frequently used in sign language. For this reason, we took 5 landmarks showing only the face and lips. For the body, we used only 6 landmarks, including shoulders and arms. Hands are the most important element in sign language, so we took all 21 poses for the hand. Thus, we used a total of 53 landmarks.

\subsubsection{\textbf{Graph Pooling}}

Graph neural networks (GNNs) use graph pooling as a technique to reduce the size of graphs while retaining important structural and contextual information. In graph pooling, the goal is to downsample or aggregate the information in a graph, similar to how pooling operations (e.g., max pooling or average pooling) are used in convolutional neural networks (CNNs) for image data.

We applied the pooling process to get the averages for every 3 frames. We applied average pooling separately for both hands, face, and body graphs.

\subsubsection{\textbf{Graph Convolution}}

GNNs use graph convolution as a technique to perform operations and feature extraction on graph-structured data. Graphs are mathematical structures made up of nodes connected by edges. We use them to represent relationships and connections between entities. In this study, we used graphs to represent the hands, face, and certain parts of the body. Before applying the graph convolution process, we combined the graphs, found separately in the form of a face, body, and hands, into a single graph by connecting the hands to the body, the body to the face, and lips.

Traditional CNNs perform convolutional operations on regular grid-like data, such as images. However, graph-structured data does not have a fixed grid structure, and nodes can have a varying number of connections. Graph convolution adapts the concept of convolution to graphs by generalizing it to work at the nodes and edges of the graph. The basic idea is to collect information from a node's neighboring nodes and update the node's representation accordingly. This process is repeated by all nodes in the graph, allowing the information to spread throughout. We also refer to this operation as message passing. The aggregation step involves combining the properties of neighboring nodes in a meaningful way. A variety of techniques can be used to achieve this, such as adding features, averaging, or applying more complex operations like weighted sums. In this study, we used the averaging method.

Graph convolutional meshes often stack multiple graph convolution layers to capture information from multiple hops in the graph. Each layer collects information from a node's local neighborhood, and as information spreads across multiple layers, it can capture more global relationships in the graph. In this study, we applied only a single graph convolution process.

\subsubsection{\textbf{Landmark Normalization}} 

For our proposed P2T-T model, we applied normalization to the hand, face, and body landmarks before sending them directly to the transformer network. By doing this normalization, we aimed to eliminate the negative effects on the model's learning process of the signer's position on the screen, the distance the signer was from the camera, and the presence of different people. To do this, we normalized the landmark coordinates we got from MediaPipe, by taking the average of the two shoulder coordinates of the signer, i.e. the midpoint of the two shoulder joints, as the origin for translation, and the distance between the two shoulder joints for the scale normalization. We applied this normalization separately for each frame. As a result, we implemented a signer centered and scale agnostic recognition process.

\subsubsection{\textbf{Transformer}}

In the context of natural language processing (NLP), transformer is a deep learning model architecture that has gained considerable popularity, particularly with the release of the groundbreaking paper "Attention Is All You Need" in 2017\cite{vaswani}.

The Transformer model replaces the traditional recurrent neural networks (RNNs) and CNNs commonly used for sequence modeling by processing sequential data such as sentences or text. At the core of the transformer model is the self-attention mechanism, also known as scaled dot product attention, or simply attention. When processing a string, self-attention allows the model to weigh the importance of different words or tokens. The correlation between each word and all the other words is extracted. In this way, the context is preserved not only for a single sentence but for the entire text. By stacking multiple layers of self-attention and feed forward neural networks, the transformer model can effectively model long-term dependencies and capture complex patterns in sequential data.

In this study, in both of our models (P2T-T and GNN-T), we used 1024 hidden units and 8 heads in each layer of our transformers. We used 6 encoder and 6 decoder layers. We set the feed forward layer dimension to 2048. These values are determined empirically.

\subsubsection{\textbf{Training}}
In all our experiments, we set the dropout ratio to 0.1, to prevent the model from overfitting. We used the Adam \cite{adam} optimizer to train our networks with a learning rate of \(5x10^{-5}\) (\(\beta_1=0.9, \beta_2=0.98\)). We utilized plateau learning rate scheduling which tracks the development set performance. If there was no improvement in the development score over 7 epochs, we reduced the learning rate by 0.7 times. We continued this process until the learning rate was \(10^{-6}\). We applied the batch size as 4 for the P2T-T model and 16 for the GNN-T model.

\subsection{\textbf{Performance Metrics}}

In this study, we used Recall-Oriented Understudy for Gisting Evaluation\cite{rouge} (ROUGE-L F-measure) and BiLingual Evaluation Understudy\cite{bleu} (BLEU) metrics to evaluate the performance of the transformer model. We use BLEU-1, BLEU-2, BLEU-3, and BLEU-4 as the BLEU score to provide a better perspective on translation performance at different expression levels, while we use the ROUGE-L F1-Score as the ROUGE score.

\begin{table*}[!htb]
\centering
\caption{Model Comparison on E-TSL Dataset}
\begin{tabular}{ p{2.5cm}|p{1.25cm}p{1cm}p{1cm}p{1cm}p{1cm}|p{1.25cm}p{1cm}p{1cm}p{1cm}p{1cm} }
 & \multicolumn{5}{c}{DEV SET} & \multicolumn{5}{c}{TEST SET} \\
 Model: & ROUGE-L & BLEU-1 & BLEU-2 & BLEU-3 & BLEU-4 & ROUGE-L & BLEU-1 & BLEU-2 & BLEU-3 & BLEU-4 \\
 \hline
 P2T-T & 22.42 & 18.26 & 8.27 & 4.82 & 3.28 & 22.09 & 18.09 & 8.20 & 4.78 & 3.23 \\
 GNN-T & \textbf{23.42} & \textbf{21.26} & \textbf{9.51} & \textbf{5.48} & \textbf{3.65} & \textbf{22.93} & \textbf{21.01} & \textbf{9.13} & \textbf{5.20} & \textbf{3.49} \\
\end{tabular}
\label{tab:model_comparison_etsl}
\end{table*}

\begin{table*}[!htb]
\centering
\caption{Model Comparison on PHOENIX14\textbf{T} Dataset}
\begin{tabular}{ p{2.5cm}|p{1.25cm}p{1cm}p{1cm}p{1cm}p{1cm}|p{1.25cm}p{1cm}p{1cm}p{1cm}p{1cm} }
 & \multicolumn{5}{c}{DEV SET} & \multicolumn{5}{c}{TEST SET} \\
 Model: & ROUGE-L & BLEU-1 & BLEU-2 & BLEU-3 & BLEU-4 & ROUGE-L & BLEU-1 & BLEU-2 & BLEU-3 & BLEU-4 \\
 \hline
 TwoStream-SLT\cite{chen} & 54.08 & 54.32 & 41.99 & 34.15 & 28.66 & 53.48 & 54.90 & 42.43 & 34.46 & 28.95 \\
 JEE-SLT\cite{camgoz2020} & - & 45.54 & 32.60 & 25.30 & 20.69 & - & 45.34 & 32.31 & 24.83 & 20.17 \\
 SMM-TLB\cite{chen2022} & 53.10 & 53.95 & 41.12 & 33.14 & 27.61 & 52.65 & 53.97 & 41.75 & 33.84 & 28.39 \\
 NSLT\cite{camgoz2018} & 31.80 & 31.87 & 19.11 & 13.16 & 9.94 & 31.80 & 32.24 & 19.03 & 12.83 & 9.58 \\
 \hline
 \hline
 Our Model (P2T-T) & 34.16 & 30.72 & 17.65 & 11.22 & 7.85 & 33.79 & 27.97 & 16.47 & 10.63 & 7.31 \\
 Our Model (GNN-T) & \textbf{34.98} & \textbf{31.76} & \textbf{18.62} & \textbf{12.38} & \textbf{9.02} & \textbf{35.46} & \textbf{32.34} & \textbf{19.28} & \textbf{12.81} & \textbf{8.93} \\
\end{tabular}
\label{tab:model_comparison_phoenix}
\end{table*}

\subsubsection{\textbf{Recall-Oriented Understudy for Gisting Evaluation (ROUGE)}}

As the name implies, automatic text summarization systems often use ROUGE as a measurement metric to evaluate their performance. Besides its widespread use in evaluating automatic text summarization systems, ROUGE also serves as a tool for evaluating machine translation. This metric measures how similar the system-generated summary or translation is to the reference summary or translation. 

ROUGE metric is generally used in three main versions, ROUGE-N, ROUGE-L, and ROUGE-W. ROUGE-N calculates n-gram similarity and is usually expressed as ROUGE-1, ROUGE-2, and ROUGE-3 (using 1, 2 and 3 as the n value). ROUGE-L measures the longest common substring (LCS) similarity. ROUGE-W, on the other hand, takes into account the sequential matching length of words.

\subsubsection{\textbf{BiLingual Evaluation Understudy (BLEU)}}

The BLEU score is a metric used to evaluate the quality of the output of machine translation systems. Introduced in a study published by IBM researchers in 2002 \cite{bleu}, BLEU measures how well a translation system's predictions match reference texts (usually created by human translators). 
The BLEU score is calculated using a method that combines the precision of n-grams, which are contiguous sequences of words from the predicted translation compared to the reference. It evaluates n-grams of various lengths, typically ranging from 1 to 4 (unigrams to four-grams), by dividing the count of n-grams in the predicted translation that match those in the reference by the total number of n-grams in the predicted translation. These precision scores for each n-gram length are weighted equally and then multiplied by a brevity penalty. This penalty adjusts for length discrepancies, penalizing translations that are shorter than their references, ensuring that the length ratio of the reference to the predicted translation does not exceed 1. The final BLEU score, therefore, reflects both the accuracy and adequacy of the translation, emphasizing translations that are both precise and appropriately verbose.

\section{\textbf{Experimental Results and Discussion}}

We conduct various experiments using the E-TSL dataset, which we created for this study to establish a baseline in sign language translation. We have categorized these experiments into two groups: P2T-T and GNN-T. The results of all experiments are presented as percentages.

Table \ref{tab:model_comparison_etsl} presents initial results from our training using two models, P2T-T and GNN-T, on the E-TSL dataset. The GNN-T model performed better than the P2T-T model in all ROUGE-L and BLEU scores. This demonstrates the success of the graph convolution process we applied in the GNN-T model. Additionally, by employing graph pooling to reduce the sequence length by a factor of three, we shortened the training duration and simplified the model's complexity. Overall, the GNN-T model not only yielded slightly better results than the P2T-T model but also offered greater ease of use.

To validate our P2T-T and GNN-T models on the newly developed E-TSL dataset, we employed the PHOENIX14\textbf{T} dataset, a well-established benchmark in sign language recognition and translation studies. The results obtained with both models are presented in Table \ref{tab:model_comparison_phoenix}. Our GNN-T model showed particularly strong performance across both datasets, achieving superior results in the ROUGE-L metric compared to the P2T-T model.

While significant progress has been made using the PHOENIX14\textbf{T} dataset, in this study, we present our initial Pose2Text results without extensively tuning the model parameters for PHOENIX14\textbf{T}. This approach allows us to assess the effectiveness of our models under less optimized conditions, using the baseline results on PHOENIX14\textbf{T} as a preliminary standard for comparison. The ROUGE-L score of our GNN-T model exceeded that of the established P2T-T model, and we also noted marginally better BLEU-1 and BLEU-2 scores. The BLEU-3 and BLEU-4 scores were very close to the baseline, demonstrating that our models are sufficiently robust for initial assessments.

\begin{table}[!htb]
\centering
\caption{Benchmarking BLEU-4 Scores}
\begin{tabular}{ p{3cm}|p{1.5cm}|p{1.5cm} }
 Dataset: & DEV SET & TEST SET \\
 \hline
 PHOENIX14\textbf{T}\cite{camgoz2018} & 9.94 & 9.58 \\
 \hline
 Elementary23-SLT\cite{voskou} & 6.67 & 5.69 \\
  \hline
 Elementary23-RAW\cite{voskou} & 0.36 & 0.00 \\
 \hline
 SWISSTXT-NEWS\cite{camgoz2021} & 0.46 & 0.41 \\
 \hline
 VRT-NEWS\cite{camgoz2021} & 0.45 & 0.36 \\
 \hline
 Our E-TSL (P2T-T) & 3.28 & 3.23 \\
 Our E-TSL (GNN-T) & 3.65 & 3.49 \\
\end{tabular}
\label{tab:dataset_comparison}
\end{table}

Table \ref{tab:dataset_comparison} displays our best BLEU-4 scores, offering a comparison of the E-TSL dataset with other datasets commonly utilized in SLR and SLT research. We also include the Elementary23-SLT dataset in our comparison due to its content similarities with ours. An initial review of the results shows that the BLEU-4 scores for the SWISSTXT-NEWS and VRT-NEWS datasets are notably lower than those for other datasets, including ours. This suggests that these benchmarks, being more general and possibly lacking in sufficient data, may not be effectively trained with state-of-the-art models, potentially limiting their reliability for future SLT studies.

The PHOENIX14\textbf{T} dataset has much narrower subject content as it consists only of weather news videos. This also allows it to have a narrower vocabulary. For this reason, the BLEU-4 score of the baseline model of the PHOENIX14\textbf{T} dataset is slightly higher than the BLEU-4 score we obtained in our dataset. Note also that data modalities of our model and the baseline model in \cite{camgoz2018} are also different.

The Elementary23-SLT dataset's baseline model results appear to be better than our dataset's baseline model results. The Elementary23-SLT dataset is actually a custom dataset derived from the much larger Elementary23\cite{voskou} dataset (approximately 71 hours). They created the Elementary23-SLT dataset by applying various methods to the Elementary23 RAW dataset, such as reducing the number of singletons, increasing the number of frequently used words, preserving various course contents, and selecting the most suitable ones for SLT. Under these settings, using the Elementary23 RAW dataset, they presented  0.36\% and 0.0\% BLEU-4 scores for the dev and test sets, respectively. With the selected SLT subset, they reported an increased BLEU-4 scores, 6.67/5.69, for the dev and test sets, respectively.  Although the vocabulary size of Elementary23-SLT dataset is slightly higher than E-TSL dataset (8.202/6.980), singleton rates (41\%/64\%) and rare word rates (75\%/85\%) are lower than E-TSL dataset. Besides, the resolution (1280x720) of the videos in the Elementary23-SLT dataset is much higher than the resolution (190x230) of the videos in our dataset. Finally, the videos in the Elementary23-SLT dataset were recorded in an optimal environment, and the signer's movements were better perceived thanks to the single-color background. All of these differences explain the disparity in performance between the baseline models of both datasets.

\section{\textbf{Conclusion}}

In this study, we introduced the E-TSL dataset, aimed at advancing the field of Continuous Sign Language Translation using Turkish Sign Language. We presented the outcomes of the Sign Pose2Text task using our newly developed P2T-T and GNN-T models on the E-TSL dataset. To validate these models, we also trained them on the PHOENIX14\textbf{T} dataset and assessed their performance. Additionally, we compared our results with those from other datasets, finding that the baseline results are promising and suggest potential for further research.

In our future work, we plan to refine the segmentation of videos in our dataset by dividing them into sentences, in addition to maintaining the current 1-minute episodes. This will allow us to create sentence-level annotations and observe how model performance varies with changes in granularity. Additionally, we intend to explore alternative pose extraction tools beyond MediaPipe and assess the impact of incorporating other state-of-the-art models on overall model performance.

\bibliographystyle{plain}
\bibliography{bib}

\begin{thebibliography}{10}

\bibitem{bragg}
D.~Bragg, O.~Koller, M.~Bellard, L.~Berke, P.~Boudrealt, A.~Braffort, N.~Caselli, M.~Huenerfauth, H.~Kacorri, T.~Verhoef, C.~Vogler, and M.R. Morris.
\newblock Sign language recognition, generation, and translation: An interdisciplinary perspective.
\newblock pages 16--31, 10 2019.

\bibitem{bull}
H.~Bull, T.~Afouras, G.~Varol, S.~Albanie, L.~Momeni, and A.~Zisserman.
\newblock Aligning subtitles in sign language videos, 2021.

\bibitem{camgoz2018}
N.~C. Camgoz, S.~Hadfield, O.~Koller, H.~Ney, and R.~Bowden.
\newblock Neural sign language translation.
\newblock In {\em 2018 IEEE/CVF Conference on Computer Vision and Pattern Recognition}, pages 7784--7793, 2018.

\bibitem{camgoz2019}
N.~C. Camgoz, S.~Hadfield, O.~Koller, H.~Ney, and R.~Bowden.
\newblock Neural machine translation for sign language: Effect of data augmentation on bleu and rouge scores.
\newblock In {\em In: Proceedings of The 16th International Conference on Machine Vision Applications (MVA’19)}, pages 1--6, 2019.

\bibitem{camgoz2020}
N.~C. Camgoz, O.~Koller, S.~Hadfield, and R.~Bowden.
\newblock Sign language transformers: Joint end-to-end sign language recognition and translation.
\newblock In {\em IEEE Conference on Computer Vision and Pattern Recognition (CVPR)}, pages 10023--10033, 2020.

\bibitem{camgoz2021}
N.~C. Camgoz, B.~Saunders, G.~Rochette, M.~Giovanelli, G.~Inches, R.~Nachtrab-Ribback, and R.~Bowden.
\newblock Content4all open research sign language translation datasets.
\newblock In {\em 2021 16th IEEE International Conference on Automatic Face and Gesture Recognition (FG 2021)}, pages 1--5, Los Alamitos, CA, USA, 2021. IEEE Computer Society.

\bibitem{chen2022}
Y.~Chen, F.~Wei, X.~Sun, Z.~Wu, and S.~Lin.
\newblock A simple multi-modality transfer learning baseline for sign language translation.
\newblock In {\em 2022 IEEE/CVF Conference on Computer Vision and Pattern Recognition (CVPR)}, pages 5110--5120, 2022.

\bibitem{chen}
Y.~Chen, R.~Zuo, F.~Wei, Y.~Wu, S.~LIU, and B.~Mak.
\newblock Two-stream network for sign language recognition and translation.
\newblock In {\em Advances in Neural Information Processing Systems}, 2022.

\bibitem{cheng}
K.~L. Cheng, Z.~Yang, Q.~Chen, and Y.W. Tai.
\newblock {\em Fully Convolutional Networks for Continuous Sign Language Recognition}, pages 697--714.
\newblock 11 2020.

\bibitem{gokul}
N.~C. Gokul, L.~Manideep, N.~Sumit, S.~Prem, K.~Pratyush, and M.~K. Mitesh.
\newblock Addressing resource scarcity across sign languages with multilingual pretraining and unified-vocabulary datasets.
\newblock In {\em Thirty-sixth Conference on Neural Information Processing Systems Datasets and Benchmarks Track}, 2022.

\bibitem{hao}
A.~Hao, Y.~Min, and X.~Chen.
\newblock Self-mutual distillation learning for continuous sign language recognition.
\newblock In {\em 2021 IEEE/CVF International Conference on Computer Vision (ICCV)}, pages 11283--11292, 2021.

\bibitem{hosain}
A.~A. Hosain, P.~S. Santhalingam, P.~Pathak, H.~Rangwala, and J.~Košecká.
\newblock Hand pose guided 3d pooling for word-level sign language recognition.
\newblock In {\em 2021 IEEE Winter Conference on Applications of Computer Vision (WACV)}, pages 3428--3438, 2021.

\bibitem{huenerfauth}
M.~Huenerfauth, L.~Zhao, E.~Gu, and J.~Allbeck.
\newblock Evaluation of american sign language generation by native asl signers.
\newblock {\em ACM Trans. Access. Comput.}, 1(1), may 2008.

\bibitem{kan}
J.~Kan, K.~Hu, M.~Hagenbuchner, A.~Tsoi, M.~Bennamoun, and Z.~Wang.
\newblock Sign language translation with hierarchical spatio-temporal graph neural network.
\newblock In {\em 2022 IEEE/CVF Winter Conference on Applications of Computer Vision (WACV)}, pages 2131--2140, Los Alamitos, CA, USA, 2022. IEEE Computer Society.

\bibitem{adam}
D.~Kingma and J.~Ba.
\newblock Adam: A method for stochastic optimization.
\newblock {\em International Conference on Learning Representations}, 2014.

\bibitem{kipp}
M.~Kipp, Q.~Nguyen, A.~Heloir, and S.~Matthes.
\newblock Assessing the deaf user perspective on sign language avatars.
\newblock In {\em The Proceedings of the 13th International ACM SIGACCESS Conference on Computers and Accessibility}, ASSETS '11, page 107–114, New York, NY, USA, 2011. Association for Computing Machinery.

\bibitem{bleu}
P.~Kishore, S.~Roukos, T.~Ward, and W.~Zhu.
\newblock Bleu: a method for automatic evaluation of machine translation.
\newblock In {\em Proceedings of the 40th Annual Meeting on Association for Computational Linguistics}, ACL '02, page 311–318, USA, 2002. Association for Computational Linguistics.

\bibitem{rouge}
C.-Y. Lin.
\newblock Rouge: A package for automatic evaluation of summaries.
\newblock pages 74--81, 2004.

\bibitem{lugaresi}
C.~Lugaresi, J.~Tang, H.~Nash, C.~McClanahan, E.~Uboweja, M.~Hays, F.~Zhang, C.~L. Chang, M.~Yong, J.~Lee, W.~T. Chang, W.~Hua, M.~Georg, and M.~Grundmann.
\newblock Mediapipe: A framework for building perception pipelines, 06 2019.

\bibitem{morissey}
S.~Morrissey and W.~Andy.
\newblock An example-based approach to translating sign language.
\newblock In {\em Machine Translation Summit}, 2005.

\bibitem{parton}
B.S. Parton.
\newblock {Sign Language Recognition and Translation: A Multidisciplined Approach From the Field of Artificial Intelligence}.
\newblock {\em The Journal of Deaf Studies and Deaf Education}, 11(1):94--101, 09 2005.

\bibitem{saunders}
B.~Saunders, N.~C. Camgoz, and R.~Bowden.
\newblock Progressive transformers for end-to-end sign language production.
\newblock In {\em Computer Vision -- ECCV 2020}, pages 687--705, Cham, 2020. Springer International Publishing.

\bibitem{Sincan20}
O.~M. Sincan and H.~Y. Keles.
\newblock Autsl: A large scale multi-modal turkish sign language dataset and baseline methods.
\newblock {\em IEEE Access}, 8:181340--181355, 2020.

\bibitem{Sincan22}
O.~M. Sincan and H.~Y. Keles.
\newblock Using motion history images with 3d convolutional networks in isolated sign language recognition.
\newblock {\em IEEE Access}, 10:18608--18618, 2022.

\bibitem{Tur19}
A.~O. Tur and H.~Y. Keles.
\newblock Isolated sign recognition with a siamese neural network of rgb and depth streams.
\newblock In {\em IEEE EUROCON 2019 -18th International Conference on Smart Technologies}, pages 1--6, 2019.

\bibitem{Tur21}
A.~O. Tur and H.~Y. Keles.
\newblock Evaluation of hidden markov models using deep cnn features in isolated sign recognition.
\newblock {\em Multimedia Tools and Applications}, 80:19137 -- 19155, 2021.

\bibitem{vaswani}
A.~Vaswani, N.~Shazeer, N.~Parmar, J.~Uszkoreit, L.~Jones, A.~N. Gomez, L.~Kaiser, and I.~Polosukhin.
\newblock Attention is all you need.
\newblock In {\em Advances in Neural Information Processing Systems}, volume~30. Curran Associates, Inc., 2017.

\bibitem{voskou}
A.~Voskou, K.~P. Panousis, H.~Partaourides, K.~Tolias, and S.~Chatzis.
\newblock A new dataset for end-to-end sign language translation: The greek elementary school dataset.
\newblock In {\em 2023 IEEE/CVF International Conference on Computer Vision Workshops (ICCVW)}, pages 1958--1967, Los Alamitos, CA, USA, oct 2023. IEEE Computer Society.

\bibitem{who}
{World Health Organization}.
\newblock Deafness and hearing loss.
\newblock Accessed: 04-13-2024.

\bibitem{Bosphorussign22k}
O.~Özdemir, A.~A. Kandıroğlu, N.~C. Camgöz, and L.~Akarun.
\newblock Bosphorussign22k sign language recognition dataset.
\newblock In {\em Proceedings of the LREC2020 9th Workshop on the Representation and Processing of Sign Languages: Sign Language Resources in the Service of the Language Community, Technological Challenges and Application Perspectives}, pages 181--188, Marseille, France, 2020. European Language Resources Association (ELRA).

\end{thebibliography}

\end{document}